\def\year{2021}\relax
\documentclass[letterpaper]{article} 
\usepackage{aaai21}  
\usepackage{times}  
\usepackage{helvet} 
\usepackage{courier}  
\usepackage[hyphens]{url}  
\usepackage{graphicx} 
\usepackage{natbib}  
\usepackage{caption} 
\usepackage{CJKutf8}
\usepackage{tabularx}
\usepackage{multirow}
\usepackage{multicol}
\usepackage{array}
\usepackage{amsmath}
\usepackage{hyperref}

\title{Social Media Reveals Urban-Rural Differences in Stress across China}
\author {
    Jesse Cui,\textsuperscript{\rm 1}
    Tingdan Zhang,\textsuperscript{\rm 2}
    Kokil Jaidka,\textsuperscript{\rm 3} 
    Dandan Pang,\textsuperscript{\rm 4} 
    Garrick Sherman,\textsuperscript{\rm 1} 
    Vinit Jakhetiya,\textsuperscript{\rm 5}
    Lyle Ungar,\textsuperscript{\rm 1}
    Sharath Chandra Guntuku\textsuperscript{\rm 1}
    \\
}
\affiliations {
    \textsuperscript{\rm 1}University of Pennsylvania, 
    \textsuperscript{\rm 2}Beijing Normal University, 
    \textsuperscript{\rm 3}National University of Singapore, 
    \textsuperscript{\rm 4}University of Bern \\
    \textsuperscript{\rm 5}Indian Institute of Technology, Jammu \\
    \{jessecui, sharathg\}@cis.upenn.edu
}
\begin{document}

\maketitle

\begin{abstract}
Modeling differential stress expressions in urban and rural regions in China can provide a better understanding of the effects of urbanization on psychological well-being in a country that has rapidly grown economically in the last two decades. This paper studies linguistic differences in the experiences and expressions of stress in urban-rural China from Weibo posts from over 65,000 users across 329 counties using hierarchical mixed-effects models. We analyzed phrases, topical themes, and psycho-linguistic word choices in Weibo posts mentioning stress to better understand appraisal differences surrounding psychological stress in urban and rural communities in China; we then compared them with large-scale polls from Gallup. After controlling for socioeconomic and gender differences, we found that rural communities tend to express stress in emotional and personal themes such as relationships, health, and opportunity while users in urban areas express stress using relative, temporal, and external themes such as work, politics, and economics. These differences exist beyond controlling for GDP and urbanization, indicating a fundamentally different lifestyle between rural and urban residents in very specific environments, arguably having different sources of stress. We found corroborative trends in physical, financial, and social wellness with urbanization in Gallup polls. 
\end{abstract}

\begin{CJK*}{UTF8}{gbsn}
\section{Introduction}
Digital trace data is increasingly used to better understand population-level health trends worldwide. An emerging body of research focuses on mining the language of self-expression on social media to understand the well-being and quality of life of the people living in cities~\citep{quercia2012talk}, regions~\citep{rentfrow2010statewide}, and countries~\citep{de2017gender}. In general this body of work has demonstrated that language expressions on social media can aid measurement of regional differences in social and cultural aspects of life.

Social media analyses on \textit{stress}, the mental or emotional strain arising from difficult circumstances, have illustrated the predictive efficacy of supervised language models trained on \textit{English} social media posts and are shown to be a valid and robust predictor of subjective mental health and well-being in communities~\citep{bagroy2017social,guntuku2019understanding,jaidka2020rural}. 

However, there are two major gaps in this body of work. First, in focusing only on social media platforms and posts from an English-speaking audience, social media activity of non-English speakers and writers is less explored. With 1.3 billion speakers or over three times as many native speakers as English, Mandarin Chinese is the most spoken language worldwide~\cite{mandarin_stats}. A relatively insular social media climate and a notable under-representation in the computational social science sphere makes China one of the, if not the most, important cultural context for understanding the relationship between social media language, mental health, and well-being in rapidly economically growing countries~\cite{jackson2013cultural}. While there are studies on language from a small number of individuals in China analyzing depression~\cite{tian2018characterizing}, suicidal ideation~\cite{wang2018study}, and anxiety~\cite{tian2017analysis}, psychological stress across regions informing public health is understudied. Second, little is understood about how the differences in social media usage relate to urban-rural experiences and socioeconomic differences. 

\begin{figure}[!t]
    \includegraphics[width=\columnwidth]{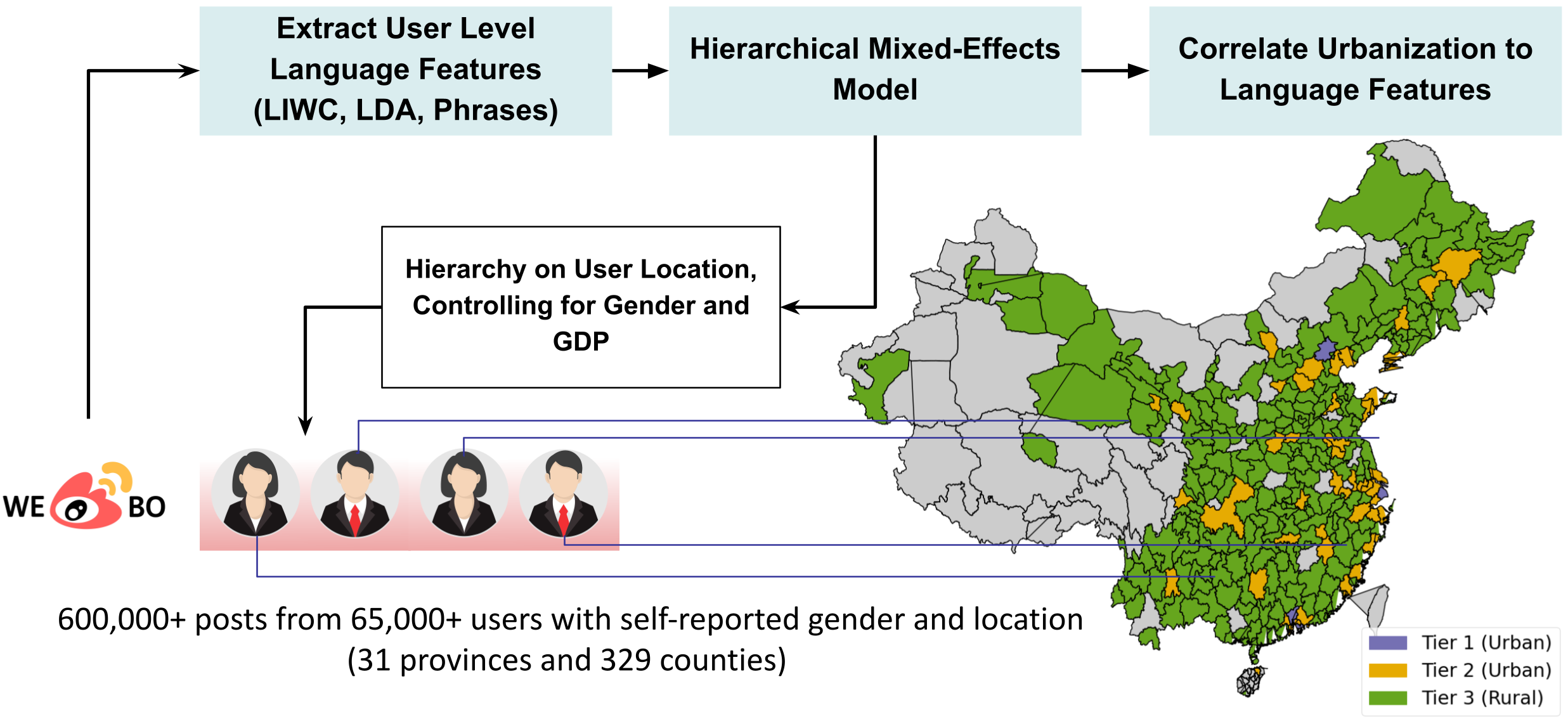}
    \caption{Hierarchical Linear Modeling to Study Urban-Rural Differences in Stress across China}
    \label{fig:model-overview}
\end{figure}

Based on a body of literature on digital inequality~\citep{dimaggio2001digital,ahmed2020internet} and the reinforcement hypothesis~\citep{robinson2015digital}, we can anticipate differences in how urban and rural residents \textit{use} social media. Therefore, digital inequality could be a factor driving regional variations in the language of social media posts. 
Analyses that recognize the heterogeneity in social media posts allow researchers to be more cognizant of demographic and social differences~\citep{gilbert10} in large scale analyses. The insights from this study can enable public health professionals to better understand nuanced differences in psychological stressors, especially in an understudied context.


Specifically, we address the following research questions in this study: 1)  How do linguistic markers of psychological stressors on Weibo differ between urban and rural Chinese communities; and 2) How do large-scale polls on comorbidities of stress compare with linguistic markers on Weibo? We answered these questions by: 1) modeling the contribution of regional characteristics on individuals' language using hierarchical linear modeling (Figure \ref{fig:model-overview}); 2) analyzing  social media users' discussion of psychological stressors in rural and urban areas in China using a) words and phrases ($n-$grams), b) topics generated from latent Dirichlet allocation (LDA), and c) psycholinguistic categories from the Chinese Linguistic Inquiry Word Count (LIWC) dictionary~\citep{zhao2016evaluating}; and 3) providing insights about the language markers of stressors on Weibo that are indicative of other comorbidities, using large scale survey data from Gallup Polls. We also discuss this study, by comparing and contrasting the findings, in the light of related work in the United States~\cite{jaidka2020rural}.  

\section{Background}
\subsection{Stress and Urbanization}


There are several potential reasons for urban and rural areas to report different well-being. For example, subjective perception of economic or class differences in urban areas lead to higher stress due to higher social comparisons~\cite{knight1999rural}. Rural communities are also known to be different than urban communities when it comes to how closely people feel connected with their family, friends, and neighbors~\cite{beaudoin2004social}. 
Since social relationships help to buffer stress and anxiety~\citep{helliwell2004social}, a lack in that domain could force an individual to cope with their stress in other ways, or experience lower well-being and higher stress as a trade-off.

Urban residents are more likely to face the environmental stressors associated with living in densely populated areas or in compact housing conditions~\citep{evans2002environment}. Major concerns to rural areas include access to healthy food and quality medical care~\cite{walker2010disparities}. On the other hand, studies have reported that an urban upbringing affects the region of the brain responsible for tasks involving negative affect processing and stress evaluation as compared to tasks involving regular cognitive processing~\citep{lederbogen2011city}.

Prior studies suggest that urban-rural differences affect personal well-being and social attitudes of their residents. This has been examined in many Western contexts, such as the United Kingdom~\citep{white2013would} and the United States~\citep{rentfrow2010statewide}. Findings from the United Kingdom~\cite{white2013would} show that individuals in urban communities with greener surroundings have lower mental distress and higher well-being, even after controlling for individual and regional covariates. Similar findings have been reported in the United States, where neuroticism is higher in the states on the east coast than on the west~\citep{rentfrow2010statewide}. Our study investigates whether, and how, these findings would generalize to the cultural and sociopolitical context of China. 

\subsection{Urban-Rural Divide in China}
Since the `Reform and Opening-up' in 1978, China's total economic output has made huge strides. However, due to urban-biased policies, the rapid growth of the Chinese economy also led to urban-rural inequality~\citep{zhu2020export}. These are reflected in almost all aspects of daily life, such as the systems of household, education, and welfare~\citep{partridge2008distance,ward2009placing}. The unique `hukou' system (i.e., house registration system) is a prominent factor for the large urban-rural divide in China as it restricts the ability of workers to move from poor rural areas to more productive urban regions~\cite{liu2005institution}. 

Compared to their rural counterparts, urban people appear to have a high level of compliance with China's one-child policy~\cite{synder2000governmental}. This leads to significantly higher ratio of boys to girls in rural areas in China. It is thus harder for rural males to find a wife. They often need to strictly follow the so-called `Caili' culture (after engagement, a man or his family has to give the bride's family a betrothal gift to get married), resulting in financial strain~\cite{jiang2015marriage}. Therefore, we hypothesized stressors related to marriage and relationships would show up in the language of rural communities.

\subsection{Weibo and Users' Traits}
Sina Weibo is a widely used platform in China, similar to Twitter. On Weibo, users can communicate, express themselves, and project their identity~\citep{bucher2017affordances}.  The language of Weibo posts thus acts as an extensive resource for deriving insights into its users' health and well-being, which is often resource intensive to obtain with traditional surveys of nationwide samples or in areas which are politically or geographically inaccessible.

While social media users are not representative of national demographics, several psychological traits can be inferred from Weibo posts, including users' age and gender~\cite{zhang2016predicting}, personality~\cite{li2014predicting}, as well as individuals' mental health~\cite{tian2018characterizing}. Weibo has also been used to study cultural differences across China and other countries~\cite{li2019exploring,guntuku2019studying,li2020studying}. Weibo has more than 550 million registered users as of 2021, 50\% of whom use it daily and more than 64,000 messages are posted every minute~\cite{weibostatista}. In this study, we used Sina Weibo posts across China to measure the linguistic differences in psychological stressors across urban and rural communities. 

\section{Variation of Weibo Stress language with Urbanization}
\subsection{Data}
Weibo posts were obtained using a breadth-first search strategy on Weibo users from prior work~\cite{guntuku2019studying}. A total of $296,932,162$ posts from $888,435$ users were collected between 2012-2019. User's self-identified profile locations were used to identify their county and province. We filtered posts by removing duplicate messages per user and only included Mandarin posts detected via \textit{langid}~\cite{lui2012langid}. We also removed users with fewer than 5 posts and only included counties with more than 100 posts. The selection of the stress keywords was determined independently by two native Chinese speakers. A joint list was generated, and differences were arbitrated. Considering the language used on social media is not standardized, relying on dictionaries was insufficient. For example, many people use ``鸭梨'' (which, literally, means white pear) to replace ``压力'' (which, literally, means stress) on social media as their pronunciations are similar and the former is netspeak. Therefore, we included words from both the thesaurus and expert curation based on different mentions of stressors on Sina Weibo. This resulted in 30 keywords. A list of the final keywords is provided in the Appendix. We sampled a random set of 100 posts with the keywords and found more than 90\% of the posts to contain mentions of stressors. This study was considered exempt by University of Pennsylvania Institutional Review Board. 

\textbf{Descriptives:} Using 30 stress-related words and phrases, we retrieved 641,262 posts from 67,027 users in 329 counties (also known as 县级市, county cities) that are spread across 22 provinces, 5 autonomous regions, and 4 municipalities. This data provided coverage for all provinces and 91\% of counties in Mainland China. Our dataset consisted of 160,064 rural messages and 481,198 urban messages from 15,916 rural users and 51,111 urban users, respectively.  29,305 users were male, 37,722 users were female. We used the three-point system for labeling counties into urban (tiers 1 and 2) and rural (tier 3) categories~\cite{chung2004china,long2016redefining}. Tier 1 counties include super-cities of China, namely Shanghai, Beijing, Guangzhou, and Shenzhen. Tier 2 counties are urbanized cities, and tier 3 counties are rural areas. We also replicated the language analyses for Weibo users at the province level, where we had access to non-binary urbanization outcome (percentage of individuals living in urban areas), and observed that the urban-rural stressor differences are consistent for both LIWC and latent Dirichlet allocation (LDA) language categories. Mandarin text along with translations for county and province level analyses are presented in Supplementary Material 1 and 2 respectively.
 
\textbf{Post-stratification:} We replicated the language analyses for Weibo users at the county level with post-stratified samples on gender, using province gender ratios from the 2015 China statistical yearbook~\cite{statistics2015china}, so that our sampled data from Weibo can match the offline population distribution~\cite{jaidka2020pnas}, and observed that the urban-rural stressor differences are consistent for both LIWC and latent Dirichlet allocation (LDA) language categories (results in Supplementary Material 3). 

\subsection{Language Features}
Weibo posts were segmented with the \textit{jieba} package in Python 3.4.
We masked specific features such as user mentions and URLs into meta tags. We aggregated posts at the user level and extracted three different features: a) words and phrases (1-3grams), b) topics generated from latent Dirichlet allocation (LDA)~\cite{blei2003latent}, and c) psycho-linguistic lexicon - Linguistic Inquiry Word Count (LIWC)~\cite{pennebaker2015development}. We used a validated simplified Chinese version of LIWC~\cite{zhao2016evaluating}. These features have been used in several social media studies on health and well-being~\cite{saha2019language,guntuku2020variability,guntuku2021social}.  

\textbf{Words and Phrases:} We used a bag of words and multi-word expressions (`phrases'), with counts normalized per user, as a linguistic dimension. These phrases ranged from one word to three consecutive words. Words and phrases that were not present in at least 10\% of the users' posts were excluded from analyses to remove outliers. Further, we use a point-wise mutual information (PMI)~\cite{bouma2009normalized}, keeping phrases with a threshold above 3 to consider only phrases that occur with high probability, e.g., so\_much\_pressure (压力\_很大). 
Word and phrase frequencies were divided by the user's total number of words, yielding relative frequencies of each.
\begin{figure*}[!t]
	\centering
	\includegraphics[width=.81\linewidth]{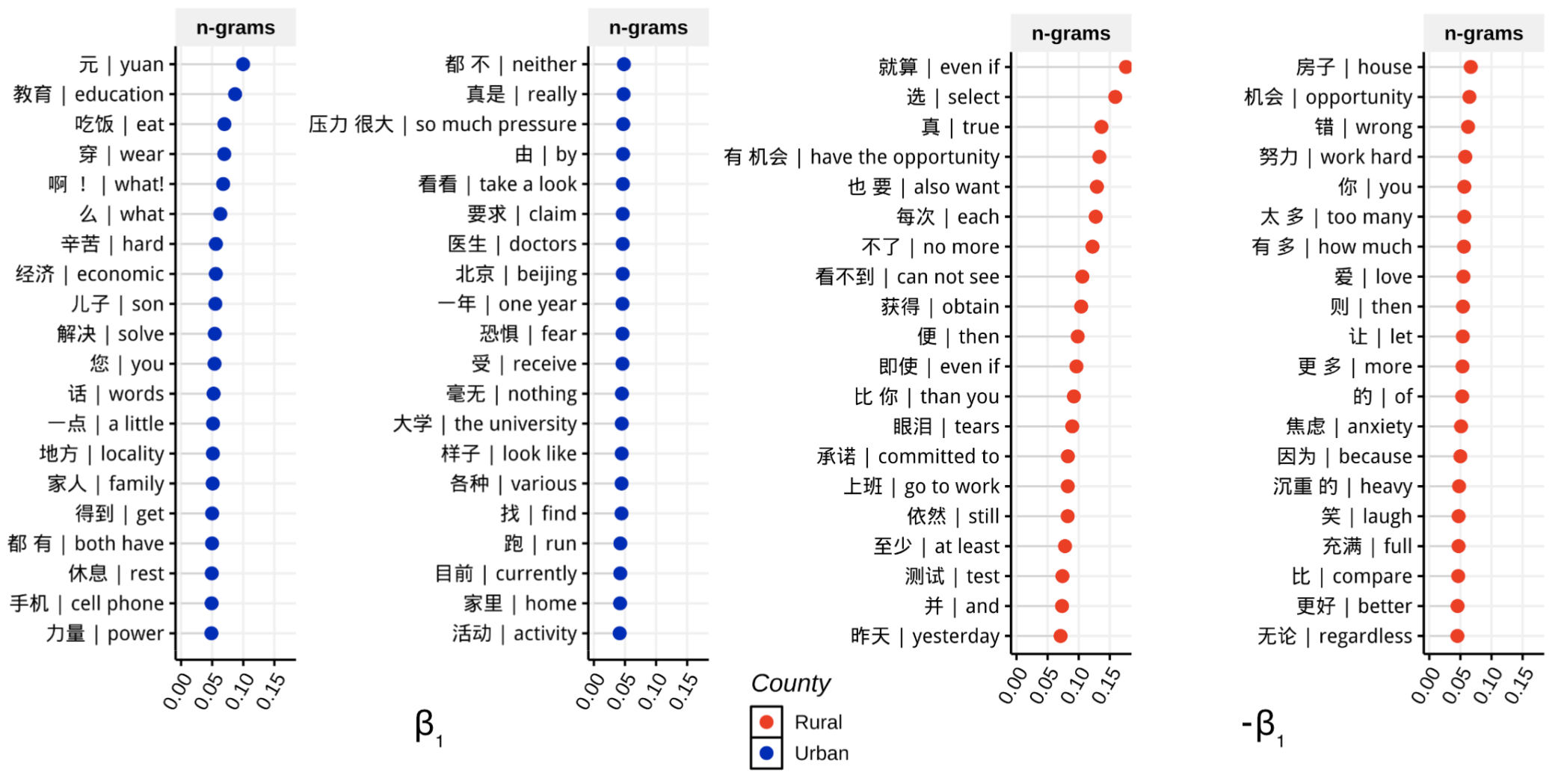}
	\caption{Rural stress phrases (in
		red) and urban stress phrases (in blue) from the top 100 most frequently occurring phrases in each area, sorted by effect size ($\beta_1$).
		Statistically significant ($p < .05$, two-tailed t-test, Benjamini-Hochberg
		corrected). Mandarin phrases and corresponding English translations are shown.}
	\label{fig:phrases}
\end{figure*}
\textbf{Topics:} We created content-specific topics limited to the thematic scope of stress expressions on Weibo using a process that consists of three steps, similar to previous works~\cite{van2020explaining,zamani2020understanding}: a) tokenization and calculating co-locations; b) identifying words most associated with stressors, and c) the topic modeling process. We used a random set of 100,000 posts each containing the stressors keywords, and another set of 100,000 random posts not containing stressors keywords. After removing re-posts and posts with URLs to minimize spam and news posts, we tokenized them and extracted the relative frequency of single words and phrases (1-3 grams) in all posts. These n-grams were used as input features in logistic regression to identify words and phrases that were significantly differentially present in tweets containing stressors keywords compared to the control set. The outcome variable was set to 1 for posts with stressors keywords and 0 for those without. To test significance, we used p \textless ~0.05 level after applying Benjamini-Hochberg correction~\cite{benjamini1995controlling}. Next, we filtered out standard stop words, the stress keywords, and any non-significant words. We then used latent Dirichlet allocation (LDA)~\cite{blei2003latent}, a probabilistic generative model which utilizes Bayesian inference to identify the prevalence of different latent linguistic states, to identify 100 topics, with an $alpha$ level of 5. We then obtained the relative topic distribution for each user. 

\textbf{LIWC:} We counted the tokens in all posts of each user that match the tokens in the LIWC dictionary based on the validated Mandarin version~\cite{zhao2016evaluating}. We then summed these token counts per user and normalized by the number of words posted by each user. 

\subsection{Hierarchical Linear Model}
An overview of the linear mixed-effects model used to analyze multi-level language data is shown in Figure \ref{fig:model-overview}. 
Level-1 of the multi-level model represents the user, the level at which we extract language features, and level-2 represents the location (county, in our case). These levels allow us to model the contribution of regional characteristics on individuals language. The use of a hierarchical mixed-effects regression modeling also allows for a robust control over location context of messages: accounting for shared variance, both within- and across-groups while not violating the independence assumption on observations~\cite{woltman2012introduction,giorgi2021well}. 

The hierarchical linear model we created treated each user-level language feature as the dependent variable and user gender and county features as the independent variable. We controlled for users' gender to account for linguistic differences with genders~\cite{schwartz2013personality} and county's log Gross Domestic Product (GDP) from the 2015 China statistical yearbook~\cite{statistics2015china} to account for its effects in relation to urbanization. The model was setup to study how being in an urban (as opposed to rural) area influences users' language attributes. We also analyzed the effect of adding a county's education level measured by the percentage of individuals who are college graduates as a control and did not find any significant difference in the results. The hierarchical model is defined in Eq. \ref{eq1}:\\
\begin{equation} \label{eq1}
\begin{split}
 Feature_i = &\beta_1*County_{IsUrban} \\
 &+\beta_2*County_{LogGDP} \\
 &+ \beta_3*User_{Gender}  + \phi +  \epsilon
\end{split}
\end{equation}

\begin{figure*}[!t]
  \centering
  \includegraphics[width=.88\linewidth]{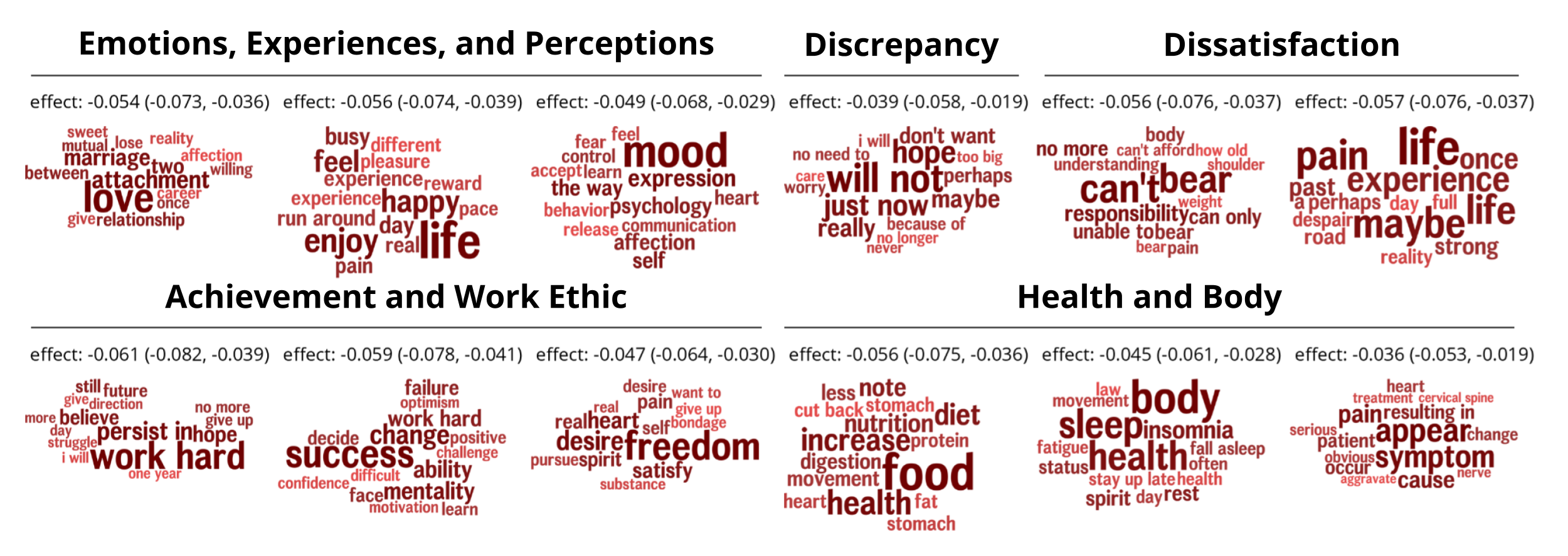}
  \caption{Topics associated with stress in rural areas. Coefficients represent the standardized effect of county urbanization to topic occurrence, significant at $p<.05$, two-tailed t-test, Benjamini-Hochberg corrected.}
  \label{fig:rural_topics}
\end{figure*}

\begin{figure*}[!t]
  \centering
  \includegraphics[width=.88\linewidth]{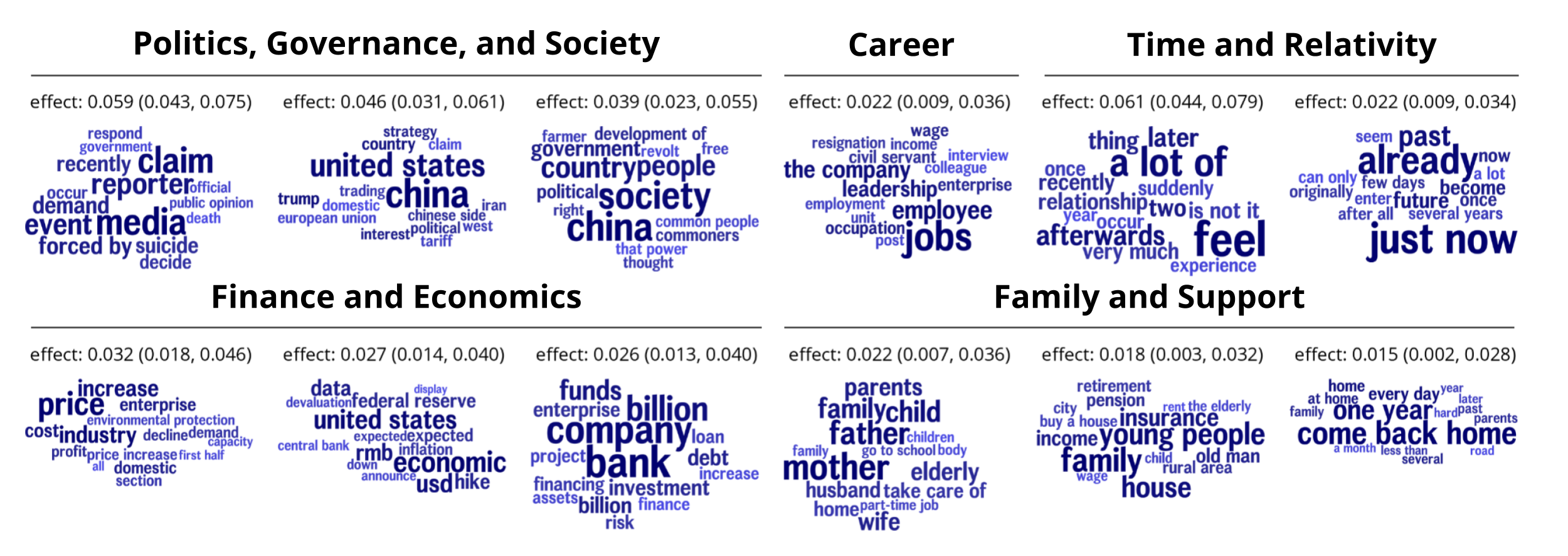}
  \caption{Topics associated with stress in urban areas. Coefficients represent the standardized effect of county urbanization to topic occurrence, significant at $p<.05$, two-tailed t-test, Benjamini-Hochberg corrected.}
  \label{fig:urban_topics}
\end{figure*}

where $\phi$ is the random effect for one of the 329 counties and $\epsilon$ is the error term. We used Benjamini-Hochberg p-correction~\cite{benjamini1995controlling} and report $\beta_1$ for all language features where correlations are significant at p$<$0.05. 

For each result, we translated Mandarin words into English for wider readability, primarily using the Google Translate Cloud API. Several words were translated by dual-lingual Mandarin and English co-authors where we believed the Google API did not accurately translate the word.

\subsection{Results}

We present and discuss language features sorted by effect size of urban-rural differences ($\beta_1$) in this section. The list of all significant language features (with Mandarin words) - a) words and phrases, b) topics, and c) LIWC categories, along with effect sizes, 95\% confidence intervals, and overall $R^2$, both conditional and marginal $R^2$, for the linear models is presented in Supplementary Material.

\textbf{Words and Phrases:}
Figure \ref{fig:phrases} shows words and phrases associated with stress in rural (red) and urban (blue) counties. 
In the context of stress mentions, individuals in the rural areas predominantly used words and phrases such as `even if', `have the opportunity', `no more', `tears', `love', `anxiety', and `work hard'  whereas, individuals in urban areas mostly used non-personal words related to society at large, such as `yuan', `education', `economic', `power', `the university', and `doctors', with the exception of family terms such as `son', `family', and `home'. 

\textbf{Topics:}
Figures \ref{fig:rural_topics} and \ref{fig:urban_topics} show rural and urban topics respectively sorted by their effect size. Rural topics surrounding stress centered around the major themes of emotions/experiences, discrepancy, dissatisfaction, achievement, and health/body. Specifically, rural stress topics contained words expressing dissatisfaction, fear, work ethic, insomnia/tiredness, health, discrepancy, power, and pursuit. Rural topics also indicate themes that are personal in nature, with many emotions, largely negative. 

Urban topics surrounding stress centered around the major themes of economics/finance, politics/society, work/careers, family/support, and time-based terms. Urban topics are predominantly around financial, economic, and market related stressors. Politics, governance, reform, career, time-based, and family support topics also surround stress in urban areas. We observed that urban stress, in contrast to rural stress, focused more on larger economic and financial issues. Also, themes around urban stress seemed to be collective, such as stress on the financial markets, stress on the economy, or stress on the people at large. 

We also pulled the messages, which have the most prevalence of a topic,  for each of the topics to observe qualitative examples. Rural messages included topics of (1) \textit{body and health}: `饮食习惯不良，使青春痘恶化，生活不 规律。 压力过大...' – translating to `Poor eating habits make acne worse; irregular life; too much stress...'; (2) \textit{love and relationships}: `经营爱情或是经营婚姻都让人疲惫 !' – translating to `Maintaining love or maintaining marriage makes people tired!'; and (3) \textit{achievement}: `压力带来动力 ，愿景决定成就' – translating to `Pressure brings motivation; vision determines achievement.'

Urban messages included stress surrounding topics of (1) \textit{career}: `
听说要绩效工资了， 上班的心情更加沉重了' – translating to `I heard that performance pay is coming, so the feeling of going to work is heavier.'; (2) \textit{finance}:`为了减轻生活压力，年轻人该如何理财' – translating to `In order to relieve the pressure of life, how should young people manage their finances?'; and (3) \textit{family}:`延迟退休是必然选择 ，有助于缓解抚养压力' – translating to `Postponement of retirement is an inevitable choice, which will help ease the pressure of supporting elderly dependency.'

\begin{table}[!t]
\begin{centering}
  \caption{LIWC categories associated with stress in rural areas; {\footnotesize significant at $p<.05$, two-tailed t-test, Benjamini-Hochberg corrected. The word 了 was untranslated as there is no corresponding English word due to its use as a perfective aspect.}}

  \label{tab:liwc_rural}
  \resizebox{\columnwidth}{!}{
  
  \begin{tabular}{|c|c|c|l|}
    \hline
    \textbf{Group} & \textbf{LIWC Category} & \textbf{Effect Size} & \textbf{Top Words (Translated)} \\
    \hline \hline
    \multirow{6}{2cm}{\centering Affective Processes} & Affect & -0.088 & pressure, correct, it is good  \\ \cline{2-4}
     & Positive emotions & -0.071 & it is good, love, happy  \\ \cline{2-4}
     & Negative emotions & -0.070 & pressure, burden, exhausted \\ \cline{2-4}
     & Sadness & -0.043 & give up, tired, pain \\ \cline{2-4}
     & Anxiety & -0.030 & pressure, upset, anxiety  \\ \cline{2-4}
     & Inhibition & -0.022 & wait, depressed, maintain  \\ \hline
    \multirow{4}{2cm}{\centering Biological Processes} & Biological processes & -0.048 & exhausted, love, eat \\ \cline{2-4}
    & Health & -0.041 & exhausted, life, health \\ \cline{2-4}
    & Sexual & -0.032 & love, woman, desire \\ \cline{2-4}
    & Body & -0.028 & feel, life, body \\ \hline
    \multirow{5}{2cm}{\centering Cognitive Processes} & Discrepancy & -0.064 & will, want, can \\ \cline{2-4}
    & Tentative & -0.027 & can, think, if \\ \cline{2-4}
    & Causation & -0.025 & let, because, so \\ \cline{2-4}
    & Cognitive mechanisms & -0.025 & 了, then, no \\ \cline{2-4}
    & Certainty & -0.019 & everyone, really, every day \\ \hline
    \multirow{5}{2cm}{\centering Function Words} & Auxiliary verbs & -0.050 & want, can, may \\ \cline{2-4}
    & Multifunction & -0.048 & of, yes, have \\ \cline{2-4}
    & Negations & -0.037 & no, none, don’t \\ \cline{2-4}
    & Conjunctions & -0.032 & then, and, also \\ \cline{2-4}
    & Quantifiers & -0.013 & many, big, more \\ \hline
    \multirow{3}{2cm}{\centering Perceptual Processes} & Perceptual processes & -0.042 & say, heavy, happy \\ \cline{2-4}
    & See & -0.034 & video, beautiful, note \\ \cline{2-4}
    & Feel & -0.024 & heavy, feel, skin \\ \hline
    \multirow{3}{2cm}{\centering Personal Concerns} & Achievement & -0.033 & jobs, need, work hard \\ \cline{2-4}
    & Home & -0.032 & family, sleep, go to bed \\ \cline{2-4}
    & Leisure & -0.028 & easy, health, drink \\ \hline
    \multirow{6}{2cm}{\centering Pronouns} & You & -0.055 & you  \\ \cline{2-4}
    & Pronouns & -0.047 & you, I, people \\ \cline{2-4}
    & Impersonal pronouns & -0.042 & people, it, that \\ \cline{2-4}
    & Personal pronouns & -0.039 & you, I, oneself \\ \cline{2-4}
    & She/he & -0.027 & he, she, people \\ \cline{2-4}
    & I & -0.018 & I, oneself, people \\ \hline
  \end{tabular}
  }
  \end{centering}
\end{table}

\begin{table}[!t]
\begin{centering}
  \caption{LIWC categories associated with stress in urban areas; {\footnotesize significant at $p<.05$, two-tailed t-test, Benjamini-Hochberg corrected. }}

  \label{tab:liwc_urban}
  \resizebox{\columnwidth}{!}{
  \begin{tabular}{|c|c|c|l|}
    \hline
    \textbf{Group} & \textbf{LIWC Category} & \textbf{Effect Size} & \textbf{Top Words (Translated)} \\
    \hline \hline
    \multirow{1}{*}{\centering Cognitive Processes} & Insight & 0.031 & 了, think, know  \\ \hline
    \multirow{1}{*}{\centering Function Words} & Specifying articles & 0.032 & on, under, before \\ \hline
    \multirow{2}{*}{\centering Informal Language} & Assent & 0.043 & 了, yes, correct \\ \cline{2-4}
    & Nonfluencies & 0.031 & just is, what, that \\ \hline
    \multirow{2}{*}{\centering Personal Concerns} & Work & 0.039 & jobs, need, work hard \\ \cline{2-4}
    & Money & 0.023 & yuan, economy, buy \\ \hline
    \multirow{2}{*}{\centering Pronouns} & They & 0.405 & they \\ \cline{2-4}
    & We & 0.026 & everyone, we \\ \hline
    \multirow{3}{*}{\centering Relativity} & Time & 0.026 & when, after, month \\ \cline{2-4}
    & Relativity & 0.018 & in, bit, to \\ \cline{2-4}
    & Motion & 0.016 & to, on, go to \\ \hline
    \multirow{1}{*}{\centering Social Processes} & Family & 0.018 & child, parents, family  \\ \hline
    \multirow{3}{*}{\centering Tenses} & Tenses & 0.064 & 了, in, ongoing  \\ \cline{2-4}
    & Progressive tense & 0.048 & 了, already \\ \cline{2-4}
    & Future tense & 0.037 & will, look, after \\ \hline
  \end{tabular}
  }
  \end{centering}
\end{table}


  

\begin{table*}[thp!]
\scriptsize
\caption{\label{tab:gallup} Regression coefficients and confidence intervals associated with Gallup well-being variables with percent urbanization. 
Coefficients are the effect size of the percent urbanization variable on the category response, and all correlations are significant at $p<.05$, two-tailed t-test. For the third question in Financial Wellness, `Living Comfortably' was taken as one class (set to 1) and remaining three were taken as another for regression (as 0).}
\begin{centering}
\begin{tabular}{|p{2cm}|p{7cm}|p{4.5cm}|p{2.5cm}|}
\hline
\textbf{Category}  & \textbf{Question} & \textbf{Responses} 
&\textbf{Effect Size (C.I.)}\\
\hline \hline
\multirow{2}{*}{General Stress}     & Did you experience the following feelings during a lot of the day yesterday? How about worry?                                                   & Yes, No                                                                         & -0.316 (-0.539,-0.093) \\ \cline{2-4}
                   & Did you experience the following feelings during a lot of the day yesterday? How about stress?                                                  & Yes, No                                                                         & -0.308 (-0.531,-0.084) \\ \hline
\multirow{2}{*}{Physical Wellness}  & Did you feel well-rested yesterday?                                                                                                             & Yes, No                                                                         & 0.177 (0.012,0.342)    \\ \cline{2-4}
                   & Did you experience the following feelings during a lot of the day yesterday? How about physical pain?                                           & Yes, No                                                                         & -0.267 (-0.416,-0.118) \\ \hline
\multirow{3}{*}{Financial Wellness} & Are you satisfied or dissatisfied with your standard of living, all the things you can buy and do?                                                & Satisfied, Dissatisfied                                                         & 0.369 (0.115,0.623)    \\ \cline{2-4}
                   & Thinking about the job situation in the city or area where you live today, would you say that it is now a good time or a bad time to find a job? & Good Time, Bad Time                                                             & 0.483 (0.249,0.717)    \\ \cline{2-4}
                   & Which one of these phrases comes closest to your own feelings about your household's income these days?                                           & [Living Comfortably], [Getting By, Finding it Difficult, Finding it Very Difficult] & 0.256 (0.140,0.373)     \\ \hline
\multirow{2}{*}{Social Wellness}    & If you were in trouble, do you have relatives or friends you can count on to help you whenever you need them, or not?                             & Yes, No                                                                         & 0.273 (0.056,0.490)     \\ \cline{2-4}
                   & In the city or area where you live, are you satisfied or dissatisfied with the opportunities to meet people and make friends?                    & Satisfied, Dissatisfied                                                         & 0.586 (0.161,1.011)    \\ \hline
\multirow{2}{*}{Civic Engagement}   & Have you done any of the following in the past month? How about donated money to a charity?                                                     & Yes, No                                                                         & 0.240 (0.052,0.427)     \\ \cline{2-4}
                   & Have you done any of the following in the past month? How about volunteered your time to an organization?                                        & Yes, No                                                                         & 0.123 (0.055,0.191)    \\ \hline
\end{tabular}
\par \end{centering}
\end{table*}

\textbf{LIWC:}
Table \ref{tab:liwc_rural} shows correlated LIWC categories for rural stress, and Table \ref{tab:liwc_urban} shows correlated LIWC categories for urban stress. 
LIWC category groups that contain categories significantly associated with rural stress are affective processes, biological processes, cognitive processes, function words, perceptual processes, personal concerns, and singular pronouns. 

Words associated with higher negative emotions and cognitive processes such as sadness, discrepancies, tentativeness, auxiliary verbs, and negations were predominant in stress mentions in rural areas. Many words in these categories suggest dissatisfaction, such as 'give up', `want', and `need'. Categories around stress in rural areas also consisted of personal pronouns and other personal constructs such as health, sexual references, achievement, home, and leisure. On the other hand, urban stress mentions were predominated by time/relativity, work/money, plural personal pronouns, and informal language. Further, while words indicative of future and progressive tenses were significantly associated with urban stress posts, words associated with past tenses were not. 

\section{Gallup Polls on Variation of Stress with Urbanization}

\subsection{Data}
To corroborate the results above with the broader societal context behind urban-rural differences in stress, we compared the insights from social media data in the context of self-reported comorbidities of stress with survey results from Gallup World Polls, which consist of survey responses from representative individuals~\cite{deaton2008income}. Annual national panels representative of the demographics in China across provinces were contacted daily by phone or face-to-face interviews by Gallup. We aggregated responses across 29 provinces and across 4 different well-being categories -- physical, social, and financial wellness along with questions measuring general stress and worry. 
We used Gallup Polls data from 2006-2011 (the entire duration for which the responses from China were available in the dataset). 17,712 responses were used in this study, averaging 2,952 responses per year across 6 years between 2006-2011. 
We then regressed the weighted averages of Gallup outcomes per province against the percent urbanization of the province, obtained from the 2015 China statistical yearbook~\cite{statistics2015china}. We specifically used, per question, the percentage of all responses that was the most affirmative response (i.e. responded "Yes" or "Satisfied") as the outcome variable for regression. Table \ref{tab:gallup} shows the questions and the available responses, as well as regression coefficients for the percent urbanization of a province. The regression model is defined in Eq. \ref{eq2} 
\begin{equation} \label{eq2}
\begin{split}
Prov._{\%Responded Yes} = \beta_1 * Prov._{\%Urban} + \epsilon
\end{split}
\end{equation}

\subsection{Differential Stress and Well-being in Urban-Rural Areas}
Rural residents reported having higher levels of personally experienced stress and worry than their urban counterparts (Table \ref{tab:gallup}). In the language analysis, we found higher occurrence of personal stress with more negative emotion words in the rural LIWC categories compared to urban language (Table \ref{tab:liwc_rural}). On the other hand, urban stress topics expanded beyond personal concerns, and the word stress itself was seen to be applied beyond the individual (Figure \ref{fig:urban_topics}).

In terms of physical wellness, urban areas were found to be more well-rested and experience less pain than their rural counterparts. The language analysis corroborated this finding, for rural users expressed more stress surrounding physical pain, tiredness, and insomnia (Figure \ref{fig:rural_topics}).

In terms of financial wellness, reports of standard of living, job opportunities, and household income comfort increased with urbanization (Table \ref{tab:gallup}). Although both urban and rural users in China expressed work stress on social media, our language analysis showed that rural users express stress surrounding (the lack of) achievement, costs, and desire, all topics that suggest dissatisfaction with one’s financial wellness (Figure \ref{fig:rural_topics}).

Lastly, we found increasing social wellness with urbanization. Urban residents indicated having reliable friends and relatives to count on in times of trouble. Our language analyses showed that both urban and rural users had stress surrounding family and relationships, but notably, topics in rural areas discussed more about personal relationships like marriage (Figure \ref{fig:rural_topics}) whereas urban topics discussed more about collective responsibilities for elders and youth (Figure~\ref{fig:urban_topics}).

\subsection{Civic Engagement}
We also used Gallup data to examine if urban users in China had more collective and societal concerns compared to rural users as we found in our language analyses that urban users expressed more stress on topics surrounding politics, governance, and society (Figure \ref{fig:urban_topics}) and used plural personal pronouns like `they' (Table \ref{tab:liwc_urban}). We observed that provinces with higher levels of urbanization also had higher levels of civic engagement. The civic engagement questions asked respondents whether they engaged in activities such as donating to charity and volunteering at organizations in their past month. People in urbanized provinces of China reported more donations to charity and more volunteering work than people in rural provinces (Table \ref{tab:gallup}). These findings are corroborated by previous studies that report that higher socioeconomic status in China is associated with higher political participation~\cite{appleton2008life}. %


\section{Discussion}
This study modelled expressions of psychological stressors in urban and rural regions in China by applying natural language processing and hierarchical mixed effects modeling to social media users across counties and provinces in China. Our findings provide a complementary perspective to better understand the urban-rural differences in stress experiences compared to traditional survey assessments~\cite{hawn2009take}. We found that rural users tended to be more personal in nature and used more emotions when discussing stress (Table \ref{tab:liwc_rural}) compared to urban users who tended to use relative and temporal terms (Table \ref{tab:liwc_urban}).

In addition, users in the rural areas had more frequent uses of singular personal pronouns, while urban users had more frequent uses of plural personal pronouns when mentioning stress. We also found that rural users appeared more concerned with personal matters, such as health, achievement, and personal relationships (Figure \ref{fig:rural_topics}), while urban users appeared more stressed about external and collective issues, such as work, politics, and economics (Figure \ref{fig:urban_topics}). These findings reflect several results obtained by previous traditional survey-based studies, in which health care, job security, and family disputes increase rural Chinese suicide risks~\cite{zhang2012factors}, lifestyle (i.e., reasonable diet, physical activity, etc.) and social support have direct or indirect effects on psychological distress in rural China~\cite{feng2013influence}, and work is the dominant stress-introducing context in urban China~\cite{lin1995urban}. We also used Gallup polls to corroborate insights from the language analysis, supporting the use of social media as a passive sensor of stressors in urban-rural areas in China.

Rural dwellers tended to directly express their symptoms of stress such as frustration, being overwhelmed, and physical pain (Figure \ref{fig:rural_topics}), while urban residents seemed more concerned with societal events that could eventually trigger stress such as the economic status of the country, the relationship between China and the US, and policies of the banks (Figure \ref{fig:urban_topics}). As the Internet is more accessible in urban than in rural areas~\cite{fong2009digital,song2020china}, urban citizens are likely more exposed to the most recent news. With higher exposure, they are more likely to talk about and be more concerned about public affairs~\cite{yao2019temporal}. These differences exist beyond controlling for GDP, indicating a fundamentally different lifestyle between rural and urban residents in very specific environments, arguably having different sources of stress~\cite{cui2012work}. 

Though rural and urban communities both seemed stressed while referring to family, topics in rural communities mentioned mainly about personal relationships like marriage (Figure \ref{fig:rural_topics}); by contrast, urban communities talked more about responsibilities for elders and youth (Figure \ref{fig:urban_topics}). This result is consistent with a previous large-scale survey which indicated that first-tier and second-tier developed cities have higher living pressure in housing, education, medical care, and supporting the elderly, while third-tier, fourth-tier, and lower-tier cities have higher levels of family and interpersonal stress~\cite{ying2016tier}. This can potentially be explained by the Confucian norms of filial piety, which `requires children to provide care for their elderly parents, putting immense social and moral pressure on adult children to fulfill their parental care responsibilities'~\cite{lee1998children,zimmer2003family}. The cost of living is higher in big cities, so the pressure to meet these obligations is greater. Moreover, urban individuals expressed more stress surrounding family (e.g., parents, retirement, illness) in comparison to the rural users (Figure \ref{fig:urban_topics}), potentially exposing more work-family conflicts in urban areas owing to more demanding jobs~\cite{ling2001work}. While rural residents could provide immediate financial, emotional, and physical care because they are more likely to live in the same household, urban residents (especially migrants) may often be absent and thus cause more stress~\cite{verheij1996explaining}. 

Prior work has shown that digital traces can predict Big-Five personality dimensions with ranging from r = .29 to .40~\cite{settanni2018predicting}, which is similar to the highest correlations between psychological traits and observable behaviors in the large psychological literature (r =~ .3). In this work, we evaluated the influence of users' communities on individuals' language and consequently expected much smaller effect sizes~\cite{hagerty2000social}. More importantly, these highly significant correlations demonstrate that Weibo posts encode significant well-being and health information that can be used to study the influence of communities in which individuals are located on their well-being.

\subsection*{Differences Between United States and China}

This study complements work which explored similar questions in the context of the United States~\citep{jaidka2020rural}. The authors used a dataset of 1.5 billion Twitter posts between January 2009 and December 2015 to identify the topics corresponding to higher stress in rural and urban United States (US). 

While there are inevitable differences in how the data in both studies were collected, prepared, and analyzed, it is nevertheless insightful to see whether the comparative differences in rural and urban areas persist across different cultural contexts. One similarity between the two studies is that higher emotional expression was associated with higher stress across rural communities, after controlling for socioeconomic status. 
The authors also reported that mentions of poor health were higher in rural areas with low socioeconomic status. We, too, found rural users to more likely discuss personal topics such as health on Weibo.

Similar to the US study, we found that discussions of politics and the economy were more likely to be posted by urban rather than rural dwellers. However, one major difference between the two studies is that, US urban communities reporting higher stress were also more likely to discuss relationships on Twitter~\citep{jaidka2020rural}. However, in our study, we did not find the same pattern. Our results revealed that urban dwellers discussed not relationships but their family as an indicator of stress (Figure \ref{fig:urban_topics}), whereas rural areas talked about marriage and romantic relationships (Figure \ref{fig:rural_topics}). This may be due to the unequal distribution of gender in China's rural areas and the so called `Caili' culture. 

\subsection*{Limitations, Ethics, and Future Work}
This study, like several social media based studies, has many limitations. First, our findings may not cover the full picture of stress because of the internet censorship in China~\cite{vuori2015lexicon} as we only use publicly available posts. We assume a few stress-related words as reported in previous research such as psychosomatic symptoms, substance use, or suicidal ideation may not be present in our dataset because they could harm the establishment of a `healthy and harmonious Internet environment'~\cite{paltemaa2020meta}. 

China's Bureau of Statistics recommends mapping neighborhood committees to urban-rural regions and we did not have access to such granular location for Weibo users in our dataset. That said, the current county tier classification system that maps counties into urban and rural regions based on a tier system serves as a reasonable proxy as has been seen in prior works~\citep{chung2004china,long2016redefining}.  Further, we replicated the findings on language data at the province-level. Discussions around the use of social media based health indices should include public health experts, computer scientists, lawyers, ethicists, clinicians, policy makers, and individuals from different socioeconomic and cultural backgrounds~\cite{benton2017ethical}.

There are several potential ways to broaden our findings. First, we analyzed posts that mention several variants of psychological stress which is potentially a subset of all posts that truly indicate a stressed mental state. While language-based estimates of stress have been validated in English using traditional survey instruments~\cite{guntuku2019understanding}, future work could examine its correlation with other well-being facets in Mandarin. Second, specific sub-populations (e.g. immigrants) who have unique stressors due to diverse reasons (e.g. `hukou') that separate rural and urban residents into disparate social, economic, and political spheres~\cite{chan2014china} could be examined. 
Despite living in cities, migrant peasant workers still maintain their rural hukou type and are treated as rural residents, with little to no access to urban social security, which can trigger different degrees and aspects of stress. 
Further, users are likely to express themselves differently across social media platforms. In this study, we only used data from one major social media platform - Weibo. It would be interesting to examine if the results would differ in other platforms, as has been found in the US~\cite{jaidka2018facebook}. 

In summary, our findings suggest that a nuanced analysis of regional social media usage is necessary, in order to situate it in an understanding of the digital divide in internet and social media usage. Differences in social contexts are related to a differential use of social media to cope with daily stressors and act as a buffer for stress and subjective well-being. Despite societal growth and technological advancements, the characteristics of urban and rural life appear to be replicated and reinforced in the online sphere -- in context after context, and culture after culture. 

\bibliographystyle{aaai21}
\bibliography{sample-base}

\begin{thebibliography}{75}
\providecommand{\natexlab}[1]{#1}
\providecommand{\url}[1]{\texttt{#1}}
\providecommand{\urlprefix}{URL }
\expandafter\ifx\csname urlstyle\endcsname\relax
  \providecommand{\doi}[1]{doi:\discretionary{}{}{}#1}\else
  \providecommand{\doi}{doi:\discretionary{}{}{}\begingroup
  \urlstyle{rm}\Url}\fi

\bibitem[{Ahmed et~al.(2020)Ahmed, Cho, Jaidka, Eichstaedt, and
  Ungar}]{ahmed2020internet}
Ahmed, S.; Cho, J.; Jaidka, K.; Eichstaedt, J.~C.; and Ungar, L.~H. 2020.
\newblock The Internet and Participation Inequality: A Multilevel Examination
  of 108 Countries.
\newblock \emph{International Journal of Communication} 14: 22.

\bibitem[{Appleton and Song(2008)}]{appleton2008life}
Appleton, S.; and Song, L. 2008.
\newblock Life satisfaction in urban China: Components and determinants.
\newblock \emph{World Development} 36(11): 2325--2340.

\bibitem[{Bagroy, Kumaraguru, and De~Choudhury(2017)}]{bagroy2017social}
Bagroy, S.; Kumaraguru, P.; and De~Choudhury, M. 2017.
\newblock A social media based index of mental well-being in college campuses.
\newblock In \emph{Proceedings of the 2017 CHI Conference on Human factors in
  Computing Systems}, 1634--1646.

\bibitem[{Beaudoin and Thorson(2004)}]{beaudoin2004social}
Beaudoin, C.~E.; and Thorson, E. 2004.
\newblock Social capital in rural and urban communities: Testing differences in
  media effects and models.
\newblock \emph{Journalism \& Mass Communication Quarterly} 81(2): 378--399.

\bibitem[{Benjamini and Hochberg(1995)}]{benjamini1995controlling}
Benjamini, Y.; and Hochberg, Y. 1995.
\newblock Controlling the false discovery rate: a practical and powerful
  approach to multiple testing.
\newblock \emph{Journal of the royal statistical society. Series B
  (Methodological)} 289--300.

\bibitem[{Benton, Coppersmith, and Dredze(2017)}]{benton2017ethical}
Benton, A.; Coppersmith, G.; and Dredze, M. 2017.
\newblock Ethical research protocols for social media health research.
\newblock In \emph{Proceedings of the First ACL Workshop on Ethics in Natural
  Language Processing}, 94--102.

\bibitem[{Blei, Ng, and Jordan(2003)}]{blei2003latent}
Blei, D.~M.; Ng, A.~Y.; and Jordan, M.~I. 2003.
\newblock Latent dirichlet allocation.
\newblock \emph{Journal of machine Learning research} 3(Jan): 993--1022.

\bibitem[{Bouma(2009)}]{bouma2009normalized}
Bouma, G. 2009.
\newblock Normalized (pointwise) mutual information in collocation extraction.
\newblock \emph{Proceedings of GSCL} 31--40.

\bibitem[{Bucher and Helmond(2017)}]{bucher2017affordances}
Bucher, T.; and Helmond, A. 2017.
\newblock The affordances of social media platforms.
\newblock \emph{The SAGE handbook of social media} .

\bibitem[{Chan(2014)}]{chan2014china}
Chan, K.~W. 2014.
\newblock China’s urbanization 2020: a new blueprint and direction.
\newblock \emph{Eurasian Geography and Economics} 55(1): 1--9.

\bibitem[{Chung and Lam(2004)}]{chung2004china}
Chung, J.~H.; and Lam, T.-c. 2004.
\newblock China's “city system” in flux: Explaining post-mao administrative
  changes.
\newblock \emph{The China Quarterly} 180: 945--964.

\bibitem[{Cui et~al.(2012)Cui, Rockett, Yang, and Cao}]{cui2012work}
Cui, X.; Rockett, I.~R.; Yang, T.; and Cao, R. 2012.
\newblock Work stress, life stress, and smoking among rural--urban migrant
  workers in China.
\newblock \emph{BMC Public Health} 12(1): 979.

\bibitem[{De~Choudhury et~al.(2017)De~Choudhury, Sharma, Logar, Eekhout, and
  Nielsen}]{de2017gender}
De~Choudhury, M.; Sharma, S.~S.; Logar, T.; Eekhout, W.; and Nielsen, R.~C.
  2017.
\newblock Gender and cross-cultural differences in social media disclosures of
  mental illness.
\newblock In \emph{Proceedings of the 2017 ACM conference on CSCW}, 353--369.

\bibitem[{Deaton(2008)}]{deaton2008income}
Deaton, A. 2008.
\newblock Income, health, and well-being around the world: Evidence from the
  Gallup World Poll.
\newblock \emph{Journal of Economic perspectives} 22(2): 53--72.

\bibitem[{DiMaggio and Hargittai(2001)}]{dimaggio2001digital}
DiMaggio, P.; and Hargittai, E. 2001.
\newblock From the ‘digital divide’to ‘digital inequality’: Studying
  Internet use as penetration increases.
\newblock \emph{Princeton: Center for Arts and Cultural Policy Studies, Woodrow
  Wilson School, Princeton University} 4(1): 4--2.

\bibitem[{Evans and English(2002)}]{evans2002environment}
Evans, G.~W.; and English, K. 2002.
\newblock The environment of poverty: Multiple stressor exposure,
  psychophysiological stress, and socioemotional adjustment.
\newblock \emph{Child development} 73(4): 1238--1248.

\bibitem[{Feng, Ji, and Xu(2013)}]{feng2013influence}
Feng, D.; Ji, L.; and Xu, L. 2013.
\newblock The influence of social support, lifestyle and functional disability
  on psychological distress in rural China: Structural equation modelling.
\newblock \emph{Australian Journal of Rural Health} 21(1): 13--19.

\bibitem[{Fong(2009)}]{fong2009digital}
Fong, M.~W. 2009.
\newblock Digital divide between urban and rural regions in China.
\newblock \emph{The Electronic Journal of Information Systems in Developing
  Countries} 36(1): 1--12.

\bibitem[{Gilbert, Karahalios, and Sandvig(2010)}]{gilbert10}
Gilbert, E.; Karahalios, K.; and Sandvig, C. 2010.
\newblock The Network in the Garden: Designing Social Media for Rural Life.
\newblock \emph{American Behavioral Scientist} 53(9): 1367--1388.

\bibitem[{Giorgi et~al.(2021)Giorgi, Guntuku, Eichstaedt, Pajot, Schwartz, and
  Ungar}]{giorgi2021well}
Giorgi, S.; Guntuku, S.~C.; Eichstaedt, J.~C.; Pajot, C.; Schwartz, H.~A.; and
  Ungar, L.~H. 2021.
\newblock Well-Being Depends on Social Comparison: Hierarchical Models of
  Twitter Language Suggest That Richer Neighbors Make You Less Happy.
\newblock In \emph{Proceedings of the International AAAI Conference on Web and
  Social Media}, volume~15, 1069--1074.

\bibitem[{Guntuku et~al.(2019{\natexlab{a}})Guntuku, Buffone, Jaidka,
  Eichstaedt, and Ungar}]{guntuku2019understanding}
Guntuku, S.~C.; Buffone, A.; Jaidka, K.; Eichstaedt, J.~C.; and Ungar, L.~H.
  2019{\natexlab{a}}.
\newblock Understanding and measuring psychological stress using social media.
\newblock In \emph{Proceedings of the ICWSM}, volume~13, 214--225.

\bibitem[{Guntuku et~al.(2021)Guntuku, Klinger, McCalpin, Ungar, Asch, and
  Merchant}]{guntuku2021social}
Guntuku, S.~C.; Klinger, E.~V.; McCalpin, H.~J.; Ungar, L.~H.; Asch, D.~A.; and
  Merchant, R.~M. 2021.
\newblock Social media language of healthcare super-utilizers.
\newblock \emph{NPJ digital medicine} 4(1): 1--6.

\bibitem[{Guntuku et~al.(2019{\natexlab{b}})Guntuku, Li, Tay, and
  Ungar}]{guntuku2019studying}
Guntuku, S.~C.; Li, M.; Tay, L.; and Ungar, L.~H. 2019{\natexlab{b}}.
\newblock Studying cultural differences in emoji usage across the east and the
  west.
\newblock In \emph{Proceedings of the International AAAI Conference on Web and
  Social Media}, volume~13, 226--235.

\bibitem[{Guntuku et~al.(2020)Guntuku, Schwartz, Kashyap, Gaulton, Stokes,
  Asch, Ungar, and Merchant}]{guntuku2020variability}
Guntuku, S.~C.; Schwartz, H.~A.; Kashyap, A.; Gaulton, J.~S.; Stokes, D.~C.;
  Asch, D.~A.; Ungar, L.~H.; and Merchant, R.~M. 2020.
\newblock Variability in language used on social media prior to hospital
  visits.
\newblock \emph{Scientific reports} 10(1): 1--9.

\bibitem[{Hagerty(2000)}]{hagerty2000social}
Hagerty, M.~R. 2000.
\newblock Social comparisons of income in one's community: Evidence from
  national surveys of income and happiness.
\newblock \emph{Journal of Personality and social Psychology} 78(4): 764.

\bibitem[{Hawn(2009)}]{hawn2009take}
Hawn, C. 2009.
\newblock Take two aspirin and tweet me in the morning: how Twitter, Facebook,
  and other social media are reshaping health care.
\newblock \emph{Health affairs} 28(2): 361--368.

\bibitem[{Helliwell and Putnam(2004)}]{helliwell2004social}
Helliwell, J.~F.; and Putnam, R.~D. 2004.
\newblock The social context of well-being.
\newblock \emph{Philosophical Transactions of the Royal Society B: Biological
  Sciences} 359(1449): 1435.

\bibitem[{Jackson and Wang(2013)}]{jackson2013cultural}
Jackson, L.~A.; and Wang, J.-L. 2013.
\newblock Cultural differences in social networking site use: A comparative
  study of China and the United States.
\newblock \emph{Computers in human behavior} 29(3): 910--921.

\bibitem[{Jaidka et~al.(2020{\natexlab{a}})Jaidka, Giorgi, Schwartz, Kern,
  Ungar, and Eichstaedt}]{jaidka2020pnas}
Jaidka, K.; Giorgi, S.; Schwartz, H.~A.; Kern, M.~L.; Ungar, L.~H.; and
  Eichstaedt, J.~C. 2020{\natexlab{a}}.
\newblock Estimating geographic subjective well-being from Twitter: a
  comparison of dictionary and data-driven language methods.
\newblock \emph{Proceedings of the National Academy of Sciences} .

\bibitem[{Jaidka, Guntuku, and Ungar(2018)}]{jaidka2018facebook}
Jaidka, K.; Guntuku, S.; and Ungar, L. 2018.
\newblock Facebook versus Twitter: Differences in self-disclosure and trait
  prediction.
\newblock In \emph{Proceedings of the International AAAI Conference on Web and
  Social Media}, volume~12.

\bibitem[{Jaidka et~al.(2020{\natexlab{b}})Jaidka, Guntuku, Lee, Luo, Buffone,
  and Ungar}]{jaidka2020rural}
Jaidka, K.; Guntuku, S.~C.; Lee, J.~H.; Luo, Z.; Buffone, A.; and Ungar, L.~H.
  2020{\natexlab{b}}.
\newblock The rural--urban stress divide: Obtaining geographical insights
  through Twitter.
\newblock \emph{Computers in Human Behavior} 106544.

\bibitem[{Jiang, Zhang, and S{\'a}nchez-Barricarte(2015)}]{jiang2015marriage}
Jiang, Q.; Zhang, Y.; and S{\'a}nchez-Barricarte, J.~J. 2015.
\newblock Marriage expenses in rural China.
\newblock \emph{The China Review} 207--236.

\bibitem[{Knight and Song(1999)}]{knight1999rural}
Knight, J.; and Song, L. 1999.
\newblock The rural-urban divide: Economic disparities and interactions in
  China.
\newblock \emph{OUP Catalogue} .

\bibitem[{Lederbogen et~al.(2011)Lederbogen, Kirsch, Haddad, Streit, Tost,
  Schuch, W{\"u}st, Pruessner, Rietschel, Deuschle et~al.}]{lederbogen2011city}
Lederbogen, F.; Kirsch, P.; Haddad, L.; Streit, F.; Tost, H.; Schuch, P.;
  W{\"u}st, S.; Pruessner, J.~C.; Rietschel, M.; Deuschle, M.; et~al. 2011.
\newblock City living and urban upbringing affect neural social stress
  processing in humans.
\newblock \emph{Nature} 474(7352): 498--501.

\bibitem[{Lee and Xiao(1998)}]{lee1998children}
Lee, Y.-J.; and Xiao, Z. 1998.
\newblock Children's support for elderly parents in urban and rural China:
  Results from a national survey.
\newblock \emph{Journal of cross-cultural gerontology} 13(1): 39--62.

\bibitem[{Li et~al.(2014)Li, Li, Hao, Guan, and Zhu}]{li2014predicting}
Li, L.; Li, A.; Hao, B.; Guan, Z.; and Zhu, T. 2014.
\newblock Predicting active users' personality based on micro-blogging
  behaviors.
\newblock \emph{PloS one} 9(1): e84997.

\bibitem[{Li et~al.(2019)Li, Guntuku, Jakhetiya, and Ungar}]{li2019exploring}
Li, M.; Guntuku, S.; Jakhetiya, V.; and Ungar, L. 2019.
\newblock Exploring (dis-) similarities in emoji-emotion association on twitter
  and weibo.
\newblock In \emph{Companion proceedings of the 2019 world wide web
  conference}, 461--467.

\bibitem[{Li et~al.(2020)Li, Hickman, Tay, Ungar, and Guntuku}]{li2020studying}
Li, M.; Hickman, L.; Tay, L.; Ungar, L.; and Guntuku, S.~C. 2020.
\newblock Studying Politeness across Cultures Using English Twitter and
  Mandarin Weibo.
\newblock \emph{Proceedings of the ACM on Human-Computer Interaction} 4(CSCW2):
  1--15.

\bibitem[{Lin and Lai(1995)}]{lin1995urban}
Lin, N.; and Lai, G. 1995.
\newblock Urban stress in China.
\newblock \emph{Social Science \& Medicine} 41(8): 1131--1145.

\bibitem[{Ling and Poweli(2001)}]{ling2001work}
Ling, Y.; and Poweli, G.~N. 2001.
\newblock Work-family conflict in contemporary China: Beyond an American-based
  model.
\newblock \emph{International Journal of Cross Cultural Management} 1(3):
  357--373.

\bibitem[{Liu(2005)}]{liu2005institution}
Liu, Z. 2005.
\newblock Institution and inequality: the hukou system in China.
\newblock \emph{Journal of comparative economics} 33(1): 133--157.

\bibitem[{Long(2016)}]{long2016redefining}
Long, Y. 2016.
\newblock Redefining Chinese city system with emerging new data.
\newblock \emph{Applied geography} 75: 36--48.

\bibitem[{Lui and Baldwin(2012)}]{lui2012langid}
Lui, M.; and Baldwin, T. 2012.
\newblock langid. py: An off-the-shelf language identification tool.
\newblock In \emph{Proceedings of the ACL 2012 system demonstrations}, 25--30.
  Association for Computational Linguistics.

\bibitem[{McCarthy(2019)}]{mandarin_stats}
McCarthy, N. 2019.
\newblock The World's Most Spoken Languages.
\newblock
  \urlprefix\url{https://www.statista.com/chart/12868/the-worlds-most-spoken-languages/}.

\bibitem[{Paltemaa et~al.(2020)Paltemaa, Vuori, Mattlin, and
  Katajisto}]{paltemaa2020meta}
Paltemaa, L.; Vuori, J.~A.; Mattlin, M.; and Katajisto, J. 2020.
\newblock Meta-information censorship and the creation of the Chinanet Bubble.
\newblock \emph{Information, Communication \& Society} .

\bibitem[{Partridge and Rickman(2008)}]{partridge2008distance}
Partridge, M.~D.; and Rickman, D.~S. 2008.
\newblock Distance from urban agglomeration economies and rural poverty.
\newblock \emph{Journal of Regional Science} 48(2): 285--310.

\bibitem[{Pennebaker et~al.(2015)Pennebaker, Boyd, Jordan, and
  Blackburn}]{pennebaker2015development}
Pennebaker, J.~W.; Boyd, R.~L.; Jordan, K.; and Blackburn, K. 2015.
\newblock The development and psychometric properties of LIWC2015.
\newblock Technical report.

\bibitem[{Quercia, Seaghdha, and Crowcroft(2012)}]{quercia2012talk}
Quercia, D.; Seaghdha, D.~O.; and Crowcroft, J. 2012.
\newblock Talk of the City: Our Tweets, Our Community Happiness.
\newblock In \emph{Proceedings of the Sixth AAAI ICWSM}, 555--558.

\bibitem[{Rentfrow(2010)}]{rentfrow2010statewide}
Rentfrow, P.~J. 2010.
\newblock Statewide differences in personality: Toward a psychological
  geography of the United States.
\newblock \emph{American Psychologist} 65(6): 548.

\bibitem[{Robinson et~al.(2015)Robinson, Cotten, Ono, Quan-Haase, Mesch, Chen,
  Schulz, Hale, and Stern}]{robinson2015digital}
Robinson, L.; Cotten, S.~R.; Ono, H.; Quan-Haase, A.; Mesch, G.; Chen, W.;
  Schulz, J.; Hale, T.~M.; and Stern, M.~J. 2015.
\newblock Digital inequalities and why they matter.
\newblock \emph{Information, communication \& society} 18(5): 569--582.

\bibitem[{Saha et~al.(2019)Saha, Kim, Reddy, Carter, Sharma, Haimson, and
  De~Choudhury}]{saha2019language}
Saha, K.; Kim, S.~C.; Reddy, M.~D.; Carter, A.~J.; Sharma, E.; Haimson, O.~L.;
  and De~Choudhury, M. 2019.
\newblock The language of LGBTQ+ minority stress experiences on social media.
\newblock \emph{Proceedings of the ACM on human-computer interaction} 3(CSCW):
  1--22.

\bibitem[{Schwartz et~al.(2013)Schwartz, Eichstaedt, Kern, Dziurzynski,
  Ramones, Agrawal, Shah, Kosinski, Stillwell, Seligman, and
  Ungar}]{schwartz2013personality}
Schwartz, H.~A.; Eichstaedt, J.~C.; Kern, M.~L.; Dziurzynski, L.; Ramones,
  S.~M.; Agrawal, M.; Shah, A.; Kosinski, M.; Stillwell, D.; Seligman, M.~E.;
  and Ungar, L.~H. 2013.
\newblock {Personality, gender, and age in the language of social media: The
  Open-Vocabulary approach}.
\newblock \emph{PLoS ONE} .

\bibitem[{Settanni, Azucar, and Marengo(2018)}]{settanni2018predicting}
Settanni, M.; Azucar, D.; and Marengo, D. 2018.
\newblock Predicting Individual Characteristics from Digital Traces on Social
  Media: A Meta-Analysis.
\newblock \emph{Cyberpsychology, Behavior, and Social Networking} 21(4):
  217--228.

\bibitem[{Song, Wang, and Bergmann(2020)}]{song2020china}
Song, Z.; Wang, C.; and Bergmann, L. 2020.
\newblock China’s prefectural digital divide: Spatial analysis and
  multivariate determinants of ICT diffusion.
\newblock \emph{International Journal of Information Management} 102072.

\bibitem[{Statistics(2015)}]{statistics2015china}
Statistics, N. 2015.
\newblock China statistical yearbook 2015.

\bibitem[{Synder(2000)}]{synder2000governmental}
Synder, M. 2000.
\newblock Governmental Control and Cultural Adaptation: A Comparison between
  Rural and Urban Reactions to China’s Fertility Control Policies.

\bibitem[{Thomala(2021)}]{weibostatista}
Thomala, L.~L. 2021.
\newblock Number of Sina Weibo users in China from 2017 to 2021
  \urlprefix\url{https://www.statista.com/statistics/941456/china-number-of-sina-weibo-users/}.

\bibitem[{Tian et~al.(2018)Tian, Batterham, Song, Yao, and
  Yu}]{tian2018characterizing}
Tian, X.; Batterham, P.; Song, S.; Yao, X.; and Yu, G. 2018.
\newblock Characterizing depression issues on Sina Weibo.
\newblock \emph{International journal of environmental research and public
  health} .

\bibitem[{Tian et~al.(2017)Tian, He, Batterham, Wang, and
  Yu}]{tian2017analysis}
Tian, X.; He, F.; Batterham, P.; Wang, Z.; and Yu, G. 2017.
\newblock An analysis of anxiety-related postings on Sina Weibo.
\newblock \emph{International journal of environmental research and public
  health} 14(7): 775.

\bibitem[{Van~Loon et~al.(2020)Van~Loon, Stewart, Waldon, Lakshmikanth, Shah,
  Guntuku, Sherman, Zou, and Eichstaedt}]{van2020explaining}
Van~Loon, A.; Stewart, S.; Waldon, B.; Lakshmikanth, S.~K.; Shah, I.; Guntuku,
  S.~C.; Sherman, G.; Zou, J.; and Eichstaedt, J. 2020.
\newblock Explaining the ‘Trump Gap’in Social Distancing Using COVID
  Discourse .

\bibitem[{Verheij(1996)}]{verheij1996explaining}
Verheij, R.~A. 1996.
\newblock Explaining urban-rural variations in health: a review of interactions
  between individual and environment.
\newblock \emph{Social science \& medicine} 42(6): 923--935.

\bibitem[{Vuori and Paltemaa(2015)}]{vuori2015lexicon}
Vuori, J.~A.; and Paltemaa, L. 2015.
\newblock The lexicon of fear: Chinese internet control practice in Sina Weibo
  microblog censorship.
\newblock \emph{Surveillance \& society} 13(3/4): 400--421.

\bibitem[{Walker, Keane, and Burke(2010)}]{walker2010disparities}
Walker, R.~E.; Keane, C.~R.; and Burke, J.~G. 2010.
\newblock Disparities and access to healthy food in the United States: A review
  of food deserts literature.
\newblock \emph{Health \& place} 16(5): 876--884.

\bibitem[{Wang et~al.(2018)Wang, Yu, Tian, Tang, and Yan}]{wang2018study}
Wang, Z.; Yu, G.; Tian, X.; Tang, J.; and Yan, X. 2018.
\newblock A study of users with suicidal ideation on Sina Weibo.
\newblock \emph{Telemedicine and e-Health} 24(9): 702--709.

\bibitem[{Ward and Brown(2009)}]{ward2009placing}
Ward, N.; and Brown, D.~L. 2009.
\newblock Placing the rural in regional development.
\newblock \emph{Regional studies} 43(10): 1237--1244.

\bibitem[{White et~al.(2013)White, Alcock, Wheeler, and
  Depledge}]{white2013would}
White, M.~P.; Alcock, I.; Wheeler, B.~W.; and Depledge, M.~H. 2013.
\newblock Would you be happier living in a greener urban area? A fixed-effects
  analysis of panel data.
\newblock \emph{Psychological science} 24(6): 920--928.

\bibitem[{Woltman et~al.(2012)Woltman, Feldstain, MacKay, and
  Rocchi}]{woltman2012introduction}
Woltman, H.; Feldstain, A.; MacKay, J.~C.; and Rocchi, M. 2012.
\newblock An introduction to hierarchical linear modeling.
\newblock \emph{Tutorials in quantitative methods for psychology} 8(1): 52--69.

\bibitem[{Yao et~al.(2019)Yao, Li, Li, and Wang}]{yao2019temporal}
Yao, L.; Li, X.; Li, Q.; and Wang, J. 2019.
\newblock Temporal and spatial changes in coupling and coordinating degree of
  new urbanization and ecological-environmental stress in China.
\newblock \emph{Sustainability} 11(4): 1171.

\bibitem[{Ying(2016)}]{ying2016tier}
Ying, X. 2016.
\newblock Residents’ sense of life pressure: a comparative study of cities.
\newblock In \emph{Annual report on social mentality of China (2016)}, 69--82.
  Social Science Literature Press.

\bibitem[{Zamani et~al.(2020)Zamani, Schwartz, Eichstaedt, Guntuku, Ganesan,
  Clouston, and Giorgi}]{zamani2020understanding}
Zamani, M.; Schwartz, H.~A.; Eichstaedt, J.; Guntuku, S.~C.; Ganesan, A.~V.;
  Clouston, S.; and Giorgi, S. 2020.
\newblock Understanding weekly COVID-19 concerns through dynamic
  content-specific LDA topic modeling.
\newblock In \emph{Proceedings of the Conference on Empirical Methods in
  Natural Language Processing. Conference on Empirical Methods in Natural
  Language Processing}, volume 2020, 193. NIH Public Access.

\bibitem[{Zhang and Wang(2012)}]{zhang2012factors}
Zhang, J.; and Wang, C. 2012.
\newblock Factors in the neighborhood as risks of suicide in rural China: a
  multilevel analysis.
\newblock \emph{Community mental health journal} 48(5): 627--633.

\bibitem[{Zhang et~al.(2016)Zhang, Caines, Alikaniotis, and
  Buttery}]{zhang2016predicting}
Zhang, W.; Caines, A.; Alikaniotis, D.; and Buttery, P. 2016.
\newblock Predicting Author Age from Weibo Microblog Posts.
\newblock In \emph{LREC}.

\bibitem[{Zhao et~al.(2016)Zhao, Jiao, Bai, and Zhu}]{zhao2016evaluating}
Zhao, N.; Jiao, D.; Bai, S.; and Zhu, T. 2016.
\newblock Evaluating the validity of simplified Chinese version of LIWC in
  detecting psychological expressions in short texts on social network
  services.
\newblock \emph{PLoS One} 11(6): e0157947.

\bibitem[{Zhu, Yu, and He(2020)}]{zhu2020export}
Zhu, S.; Yu, C.; and He, C. 2020.
\newblock Export structures, income inequality and urban-rural divide in China.
\newblock \emph{Applied Geography} 115: 102150.

\bibitem[{Zimmer and Kwong(2003)}]{zimmer2003family}
Zimmer, Z.; and Kwong, J. 2003.
\newblock Family size and support of older adults in urban and rural China:
  Current effects and future implications.
\newblock \emph{Demography} 40(1): 23--44.

\end{thebibliography}

\appendix
\section{Appendix}
\textbf{Stress Keywords} along with literal translations:
{\footnotesize 压力 | Pressure, 
鸭梨 | Pear, 
压迫 | Oppression, 
负担 | Burden, 
累赘 | Burden, 
重负 | Heavy burden, 
重担 | Heavy burden, 
重压 | Heavy pressure, 
压抑 | Suppress, 
沉重 | Heavy, 
压力大 | High pressure, 
压力山大 | Tremendous stress, 
心理压力 | Psychological stress, 
慢性压力 | Chronic stress, 
生活压力 | Life stress, 
工作压力 | Work stress, 
经济压力 | Economic stress, 
精神压力 | Mental stress, 
婚姻压力 | Marriage stress, 
环境压力 | Environmental pressure, 
租房压力 | Rental pressure, 
买房压力 | Pressure of buying a house, 
职场压力 | Career stress, 
就业压力 | Employment stress, 
年龄压力 | Age stress, 
颜值压力 | Appearance stress, 
同辈压力 | Peer pressure, 
学习压力 | Study-induced stress, 
学业压力 | Academic stress and, 
考试压力 | Exam stress.

Posts containing certain keywords associated with market stress, physical pressure, etc. were dropped to reduce false positives; specifically, 资本压力 | Capital stress, 胎压 | Tire pressure, 轮胎压力 | Tire pressure, 股票 | Stock, 大盘 | Market index, 指数 | Index, 压力位 | Resistance level, 压力线 | Resistance line, 支撑位 | Support level, 支撑线 | Support line, 压迫力 | Constriction, 承压 | Compression, 交通压力 | Traffic pressure, 企业压力 | Enterprise pressure, 肾脏负担 | Kidney pressure and, 皮肤压力 | Skin pressure. 
}
\\
\\
\noindent \textbf{Post-Stratification:}
We also replicated the language analyses for Weibo users at the county level with post-stratified samples on gender so that our sampled data from Weibo can match offline population gender ratios per province (results shown in Supplementary Material 3). 

We reweighed each language feature as defined in Eq. \ref{eq4}:\\

\begin{equation} \label{eq4}
\footnotesize
\begin{split}
PoststratFactor_p &= \frac{WeiboMale_p}{WeiboFemale_p} * \frac{PopFemale_p}{PopMale_p} \\
Feature_{i_{poststratified}} &=
\begin{cases}
Feature_{i} * PoststratFactor_p \\ \qquad \qquad \qquad \qquad  \text{ if user is female} \\
Feature_{i} * \frac{1}{PoststratFactor_p} \\ \qquad \qquad \qquad \qquad \text{ if user is male}
\end{cases}
\end{split}
\end{equation}
where $p$ indicates the province that the user is located in.
\\
\\
\noindent \textbf{Code} and Supplementary Material at \url{https://github.com/jessecui/WWBP-China-Urban-Rural-Stress}

\end{CJK*}

\end{document}